\documentclass{SCIS2025}

\usepackage{graphicx}

\begin{document}
\ArticleType{POSITION PAPER}
\Year{2025}
\Month{}
\Vol{}
\No{}
\DOI{}
\ArtNo{}
\ReceiveDate{}
\ReviseDate{}
\AcceptDate{}
\OnlineDate{}
\AuthorMark{}
\AuthorCitation{}

\title{The Superalignment of Superhuman Intelligence with Large Language Models}

\author[1]{Minlie Huang}{aihuang@tsinghua.edu.cn, Minlie Huang}
\author[1]{Yingkang Wang}{}
\author[1]{Shiyao Cui}{}
\author[2]{Pei Ke}{}
\author[3]{Jie Tang}{}


\address[1]{The CoAI group, DCST, Tsinghua University, Beijing, 100084, China}
\address[2]{University of Electronic Science and Technology of China,  611731, China}
\address[3]{The Knowledge Engineering Group (KEG), Tsinghua University, Beijing, 100084, China}

\abstract{We have witnessed superhuman intelligence thanks to the fast development of large language models and multimodal language models. As the application of such superhuman models becomes more and more popular,  a critical question arises here: how can we ensure superhuman models are still safe, reliable and aligned well to human values?
In this position paper, we discuss the concept of superalignment from the learning perspective to answer this question by outlining the learning paradigm shift from large-scale pretraining, supervised fine-tuning, to alignment training. We define superalignment as designing effective and efficient alignment algorithms to learn from \textit{noisy-labeled} data (point-wise samples or pair-wise preference data) in a \textit{scalable} way when the task becomes very complex for human experts to annotate and the model is stronger than human experts.
We highlight some key research problems in superalignment, namely, weak-to-strong generalization, scalable oversight, and evaluation. We then present a conceptual framework for superalignment, which consists of three modules: an \textit{attacker} which generates adversary queries trying to expose the weaknesses of a learner model; a \textit{learner} which will refine itself by learning from scalable feedbacks generated by a critic model along with minimal human experts; and a \textit{critic} which generates critics or explanations for a given query-response pair, with a target of improving the learner by criticizing. 
We discuss some important research problems in each component of this framework and highlight some interesting research ideas that are closely related to our proposed framework, for instance, self-alignment, self-play, self-refinement, and more. Last, we highlight some future research directions for superalignment, including identification of new emergent risks and multi-dimensional alignment. 
}


\maketitle

\section{Introduction }


The fast development of generative AI, typically known as Large Language Models (LLM) or Multimodal Language Models (MLM)\footnote{In this paper, we will only focus on LLMs, but many claims are also applicable to MLMs.}, has garnered significant attention due to its emerging ability to tackle a large variety of complex tasks, including mathematics, reasoning, coding, visual understanding and generation, and social tasks~\cite{openai-chatgpt, zhao-2023-survey}. These models have shown unbelievable competence and demonstrated human-level or even beyond-human-level performance on many benchmarks. 
This progress has fueled discussions about the concept of superhuman intelligence or Artificial General Intelligence (AGI) whose definition has not been widely accepted. To name a few definitions, OpenAI defines AGI as highly autonomous systems that outperform humans at most economically valuable work~\cite{openai-charter}. While Gary Marcus, a cognitive scientist from New York University, defines AGI as any intelligence that is flexible and general, with resourcefulness and reliability comparable to or beyond human intelligence\cite{gary-2022-agi}. However, there are also debating opinions, as Yann Lecun said, ``Human intelligence is NOT general", what we are discussing is actually advanced machine intelligence (AMI)\cite{lecun-2024-human,lecun-2024-ami}. 
As a pivotal milestone in artificial intelligence research, AGI aspires to emulate human-like cognitive versatility, enabling it to reason, make decisions, and solve problems in dynamic, unpredictable environments, with very generalizable manners.

Along with their tremendous capabilities, these superhuman models also raise critical ethical, safety, and governance concerns which may pose severe threats to human society~\cite{Shevlane-2023-risks}.
%
%
In particular, highly intelligent models possess a greater capacity for autonomous decision-making, making them harder to predict and control. 
This raises significant concerns about unintended behaviors, especially in high-stakes applications such as finance, healthcare, and critical infrastructure~\cite {bengio-2024-risks}.
As unaligned LLMs could pose substantial risks to humanity~\cite{Shen-2023-alignment}, great efforts have been made to align LLMs with human values through learning from human feedback using alignment algorithms such as PPO\cite{John-2017-ppo}, DPO (direct preference optimization)\cite{Rafael-2023-dpo}, EXO (efficient and exact alignment optimization)\cite{ji-2024-exo}, and many more.
These models, after being pretrained on large-scale corpora, are aligned with well-curated human preference data which manifest human values, social norms, or ethical concerns, either implicitly or explicitly. 
%
%


Since the capabilities of LLMs have grown very fast and we have already witnessed superhuman intelligence in many tasks, critical questions arise: \textit{how can we ensure these systems remain safe and aligned well with human values, and how can we control the behaviors of such superhuman systems?} This concern grows even more serious with the potential development of superhuman intelligence, namely AI systems that exceed human intelligence in nearly all domains~\cite{openai-sa} and maintain the ability to self-evolve its abilities automatically.
In this setting, vanilla alignment techniques relying on human feedback will not be applicable anymore since the tasks are becoming more and more complex, and the systems are even more intelligent than our humans so that even human experts cannot provide scalable and reliable supervision to supervise the learning of superhuman AI systems. In other words, traditional alignment algorithms and human supervision cannot scale further when 1) the task becomes extremely difficult (e.g., olympic competition-level coding), and 2) the system intelligence is beyond even human experts. 

Therefore, to ensure the safety of superhuman models necessitates superalignment\footnote{Superalignment was firstly introduced by OpenAI, however, we will state what the term exactly means in this paper.}, which seeks to automatically align these superhuman models with human values by means of scalable, reliable, and generalizable manners. 
Superalignment facilitates alignments via self-refinement or self-play driven by interactions and collaboration among AI models. 
Unlike traditional alignment methods, humans in superalignment only play a minimal role in assisting the automatic alignment process, where the alignment is realized through a ``human-in-the-loop" paradigm: the superhuman model is learned from automatically scalable feedbacks and human experts only provide supervision on a small proportion of cases. 

This paper is structured as follows: in Section 2, we give a definition to superalignment from the machine learning perspective, where we address the typical learning paradigms of large-scale pretraining, supervised fine-tuning, and LLM's alignment; in Section 3, we highlight some key research problems in superalignment including weak-to-strong generalization, scalable oversight, and evaluation; in Section 4, we present a feasible framework for superalignment which consists of three modules, namely attacker, learner and critic, and discuss critical research issues in each module and some interesting attempts to this framework; finally, we summarize this paper and also highlight some important future directions.

\section{Definition of Superalignment from the Learning Perspective}

In this section, we will formally introduce the concept of superalignment. We will start from the learning paradigm of large-scale pretraining, then introduce classical alignment algorithms of large language models, and finally describe the meanings of superalignment from the machine learning perspective.
The learning processes of a powerful LLM fall into three major steps: pretraining from trillions of unlabeled data, supervised fine-tuning on human-curated query-response pairs, and alignment from human preference data.

\subsection{Learning paradigm of large-scale pretraining} 

During pretraining, a model learns a generation distribution  $P_\theta$ from large-scale text corpora $\mathcal{D}$ sampled from an unknown, underlying data distribution $P_{data}$ , which is well-known as the \textit{next-token-prediction} learning paradigm:

\begin{equation}
   \mathcal{L}_{\mathrm{pretraining}} = -\mathbb{E}_{x \sim P_{data}(x)}  \sum_{i=1}^{T}{{\rm{log}} P_{\theta}(x_i | x_{<i}) }
\end{equation}
where each $x$ means a text segment consisting of $T$ tokens, each  $x_i$ denotes a token, and $x_{<i}$ indicates the preceding context of $x_i$. Given a huge amount of text, the model will learn the generation distribution $P_{\theta}(x)$ in an unsupervised way.  However, next-token-prediction can date back to 2014 since neural generation models \cite{Dzmitry-2015-nmt} have been used for machine translation or other sequence-to-sequence transformation tasks. In such a framework, the model tries to translate a source sequence $x$ to a target sequence $y$ by generating target tokens in an autoregressive way, as follows:

\begin{equation}
   \mathcal{L}_{\mathrm{sft}} = -\mathbb{E}_{(x,y) \sim P_{data}(x,y)}  \sum_{i=1}^{T}{{\rm{log}} P_{\theta}(y_i | y_{<i}, x) }
   \label{eq:sft}
\end{equation}

The model is trained on a corpus of $(x,y)$ pairs, where the supervision signal is derived from the target sequence $y$, either constructed by human annotation or automatically from unsupervised data.

\subsection{Alignment training of large-scale language models} 

A pre-trained model can demonstrate surprisingly good cross-task, few shot generalization performance, however, it is still not sufficient for generating results that are well aligned with human values. Therefore, alignment training is crucial for improvement, where the model will be further trained on a dataset consisting of high-quality human-curated $(x,y)$ pairs where human values are implicitly or explicitly embedded in the data. The training objective is the same as that in Eq. \ref{eq:sft}, where this process is usually named as \textit{supervised fine-tuning (SFT)}. The construction of the data pairs normally considers human values such as safety issues, social norms, and ethical concerns. 

During supervised fine-tuning, we only teach the model to learn what is a good generation, namely, negative examples are not used for learning. However, in the human learning process, we are always learning from both positive and negative examples.
Thus, we can learn from paired preference data by constructing data triples $D=\{(x,y_w,y_l)\}$, where for a given input $x$, a winning response  $y_w$ with higher quality and a loss response $y_l$ with lower quality is built. On top of such data triples, we can first learn a reward function $r(x,y)$ which rates how well a response $y$ can respond to an input query, and then apply some alignment algorithm to learn from such preference data.

The most popular and effective alignment algorithm is reinforcement learning from human feedback with PPO \cite{John-2017-ppo}. The learning objective is presented as follows: 

\begin{equation}
   \mathcal{L}_{\mathrm{RLHF}}=-\mathbb{E}_{x \sim P_{data}(x,y)}\left[\mathbb{E}_{y \sim P_\theta(y \mid x)}[r(x, y)]-\beta \mathbb{D}_{\mathrm{KL}}\left(P_\theta(y \mid x) \| P_{\mathrm{sft}}(y \mid x)\right)\right]
   \label{eq:ppo}
\end{equation}
where $P_\mathrm{sft}$ is the generation distribution obtained via supervised fine-tuning, $P_{\theta}$ is the distribution to be optimized during alignment, $\mathbb{D}_{\mathrm{KL}}(p||q)$ is the KL divergence between two distributions $p$ and $q$, and $\beta$ is a hyperparameter weighting the regularization term. 

The PPO algorithm has been shown very effective and widely used in alignning a pretrained LLM. However, it becomes very slow since it requires online sampling during the training process when the model size and training data are large. Thus, several methods are proposed to stabilize and accelerate the training process by avoiding reinforcement learning. For example, direct preference optimization (DPO) \cite{Rafael-2023-dpo} extracts the optimal policy from the standard RLHF objective in a closed form, thereby solving RLHF with a simple classification loss:

\begin{equation}
    \mathcal{L}_{\mathrm{DPO}}=-\mathbb{E}_{(x,y_w,y_l)\sim P_{data}(x,y_w,y_l)}\left[\log \sigma\left(\beta \log \frac{P_{\theta}(y_w|x)}{P_{\mathrm{ref}}(y_w|x)}-\beta \log \frac{P_{\theta}(y_l|x)}{P_{\mathrm{ref}}(y_l|x)}\right)\right]
\end{equation}
where $P_{\mathrm{ref}}(.|.)$ is usually a generation distribution obtained via supervised fine-tuning.

Essentially, DPO is a maximum likelihood estimation method, where the learning objective tries to increase the likelihood of observing the winning response and yet decrease that of observing the loss response \cite{Rafael-2023-dpo}. Due to its simplicity and effectiveness, DPO has become popular and many variants have been proposed, which mainly fall into three types: first, leverage preference from single human reference \cite{yuanself};
second, change the preference data distribution using rejection sampling \cite{liu2023statistical}, or extend pair-wise preference to ranking preference data \cite{song2024preference};
third, modify the learning objective such as maximizing human utility based on prospect theory \cite{ethayarajh2024kto} or substituting the point-wise reward with a pair-wise preference function \cite{azar2024general}. 



\subsection{Superalignment of large-scale language models} 
In alignment training of LLMs, there are underlying assumptions that are usually neglected. As shown in Eq. \ref{eq:sft} (next-token-prediction), we actually assume that the next token is a golden target without any noise during the supervised fine-tuning phase. In Eq. \ref{eq:ppo}, we implicitly assume that the reward model for rating a query-response pair $(x,y)$ is learned from perfect human preference data and we can learn a perfect reward function. However, in superalignment, these assumptions do not hold any more because of two facts: first, the task itself becomes very complex such that even human experts cannot provide reliable annotations, thereby leading to noisy human labels; second, the model becomes super intelligent and is even smarter than our humans, thus human experts cannot identify the flaws of a generated response, or reliably distinguish the quality difference between two responses. In other words, during superalignment, we only have noisy labels/annotations for training a superhuman model during both supervised fine-tuning (e.g., as in Eq. \ref{eq:sft}) and learning from human feedback (e.g., as in Eq. \ref{eq:ppo}). 
 
Now, let us come to the definition of superalignment from the learning perspective: superalignment is about designing effective and efficient alignment algorithms to learn from \textit{noisy-labeled} data (point-wise samples or pair-wise preference data) in a \textit{scalable} way when the task becomes very complex for human experts to annotate and the model is stronger than human experts. The superalignment setting raises some fundamental research problems which will be detailed in the next section. 

\section{Key Research Problems in Superalignment}
There are fundamental research problems in superalignment. These problems are closely related to answering these questions: how can we continuously improve a superhuman model that is even more intelligent than our humans, and how can we ensure the superhuman model is still controllable, safe, and well-aligned with human values?

More specifically, we will discuss the below research problems in the following sections:
\begin{itemize}
    \item \textbf{Weak-to-strong generalization}: how to align and improve strong models with weak supervisors? In this setting, a stronger model is supervised by a weaker model or a human (weaker than the strong model in superalignment) but we are seeking to align and further improve the stronger model. 

    \item \textbf{Scalable oversight}: how to provide scalable and reliable supervision signals to train strong models from human or AI models when the task is overly complex or even human experts cannot make reliable annotations.  

    \item \textbf{Evaluation}: how to validate the alignment of superhuman models by automatically searching for problematic behaviors and problematic internals, and how to conduct adversarial tests automatically to expose the weaknesses of strong models?

\end{itemize}

\subsection{Weak-to-strong generalization} 
Weak-to-strong generalization aims to optimize a stronger model continuously using a weaker supervisor, which was first introduced by OpenAI\cite{openai-2024-w2s}. In traditional machine learning tasks, a target model (to be optimized) is weaker than the supervisor which is usually human, or a stronger model (well known as knowledge distillation). However, in superalignment, the supervisor is even weaker than the superhuman model to be optimized, which poses new challenges to further improve the superhuman model. 

OpenAI made an analogy to this setting\cite{openai-2024-w2s}. They supervised GPT-4 with a GPT-2-level model on NLP tasks, and they define \textit{performance gap recovered (PGR)} which measures the fraction of the performance gap between the weak and strong ceiling models that we can recover with weak supervision. They found that the resulting model typically performs somewhere between GPT-3 and GPT-3.5. In this manner, they were able to recover much of GPT-4’s capabilities with only much weaker supervision. 
This research manifests that a strong model can generalize beyond weak supervision, solving even hard problems for which the weak supervisor can only give incomplete or flawed training labels. 

There are some important research sub-problems in weak-to-strong generalization. Since OpenAI's study is still very preliminary, there is yet much space to explore in this direction. First, since the supervisor is weaker, which information will be useful for supervising the stronger model and how to identify such information? Second, since the supervision signal is noisy, how can the stronger model learn robustly from noise samples?  This problem has been studied extensively in machine learning communities, however, it becomes much more complex in The setting of LLMs as the noises may be imposed at the token, span, or response level and the generative learning problem is more difficult than simple classification or regression problems. 
Third, since in general purpose a stronger model is very hard to learn from weaker supervision, can we assemble multiple specialized weaker models to supervise the learning of a stronger model?


\subsection{Scalable oversight}

Scalable oversight aims to empower relatively weak overseers to deliver reliable supervision, including training labels, reward signals, or feedback~\cite{Samuel-2022-scalable}, for complex tasks. As superalignment needs to tackle extremely complex and highly intelligent AI models,  scalable oversight can provide a technical road to overcome the limitations of human supervision, providing reliable oversight of great quality. There are two feasible paths towards providing scalable oversights: one is to use powerful models to provide scalable feedbacks and the other is to assist human annotators with strong critic models so that humans can easily provide supervision on complex tasks.

Existing proposals for scalable oversight mainly fall into three types. The first type is about decomposition. Task decomposition is a representative paradigm to provide scalable oversight, where the complex task is decomposed into a series of relatively simpler subtasks that can be more easily handled. For instance, iterated amplification~\cite{Paul-2018-IA} constructs training signals iteratively by integrating solutions to simpler subtasks. Wen et al.\cite{Wen-2024-programming} demonstrate that competition-level code generation can be solved more efficiently by decomposing a complex program into sub-functions, which they called human-centric decomposition.  Similarly, recursive reward modeling~\cite{Jan-2018-rrm} enhances AI models by progressively supervising them using reward models that are iteratively refined through improved human feedback.
The second type utilizes a powerful model to generate feedback, critiques, and labels in accordance with human-designed principles to acquire scalable oversight~\cite{Sun-2023-Principle}. Anthropic applies this approach during the reinforcement learning (RL) phase with a trained preference model to provide rewards~\cite{Bai-2022-Constitutional}, marking a shift from ``Reinforcement Learning from Human Feedback" (RLHF) to ``Reinforcement Learning from AI Feedback" (RLAIF).
In the third type, scalable oversight can be achieved via debate between multiple AI agents to determine the best answer to a given question~\cite{Du-2024-debate,Geoffrey-2018-debate}. During the process, humans play a minimal role by providing the necessary rules to guide the debate and acting as the final arbiter to select the most appropriate response.

Despite the efforts above, there are key research problems unsolved in scalable oversight. First, can we build a universal model to provide critics or feedbacks in a scalable and generalizable manner, which works for all tasks and settings? Though GPT-4 has shown very general critic ability for all types of tasks, how such ability is acquired is still unclear. Second, how can human experts be assisted by a copilot model (e.g., CriticGPT\cite{critiquegpt}) to provide reliable feedback or annotation for extremely challenging tasks? Third, how can human and AI models collaborate together to provide scalable feedback for superalignment?


\subsection{Evaluation} 

Evaluation aims to measure the alignment of superhuman models accurately from different dimensions and automatically reveal the weaknesses of superhuman models. Although evaluation has been a long-standing research problem in NLP, existing evaluation metrics cannot reflect the quality of generated texts from superhuman models since their performance has surpassed humans, which poses severe challenges to this important constituents of superalignment.

Existing works on evaluation for alignment of AI models fall into three categories: 
1) Benchmarks: Most of the existing benchmark datasets aim to measure specific abilities of LLMs on fixed benchmark datasets, including math \cite{cobbe2021training}, reasoning \cite{hendrycks2021mmlu}, code generation \cite{chen2021evaluating}, and instruction following \cite{llmjudge}. However, these static benchmark datasets face severe challenges in data pollution, thereby causing over-estimated performance especially on subsequent LLMs that may use similar data as training data. 
Thus, considering the evaluation of superalignment of AI models, the benchmark dataset should be constructed dynamically and updated quickly by including high-quality and diverse samples which can consistently reveal the weaknesses of fast-growing superhuman models.
2) LLM-based evaluation method: Existing works mostly utilize the ability of current LLMs to measure the generation quality \cite{llmjudge}. Specifically, they
formulate evaluation as an instruction-following QA task, and use LLMs to generate both evaluation scores and explanations via elaborate prompt design \cite{llmjudge}. The ability to generate evaluation results in an unsupervised manner may come from the pre-training data which are similar to the evaluation task such as comments and reviews. To automatically evaluate the alignment of superhuman models, it is important to trace the root of the evaluation ability of AI models and thus fully stimulate this ability for generation quality assessment.
3) Critic model: 
To achieve superalignment of AI models, it is important to construct a universal critic model which can efficiently provide evaluation results in a large variety of tasks and settings \cite{Ke-2024-Critique,Nat-2024-criticGPT}. Such critic model can provide scalable feedback to further improve AI models in various tasks, thereby assisting the superalignment of AI models. Existing works have also connected critique generation with reward models \cite{zhang2024generative}, which indicates a promising way to collect high-quality reward signals for guiding superhuman models towards stronger generation capabilities.


Despite the rapid development of evaluation, there still exist some essential research problems towards superalignment. First, how to automatically construct adversarial datasets to expose the weaknesses of superhuman models? This problem is under-explored because most of the existing benchmarks are restricted to human-crafted task taxonomies, thereby only revealing the weaknesses in these tasks. Some preliminary studies have shown that well-designed pipelines based on state-of-the-art LLMs (such as GPT-4) can automatically find the weaknesses in LLMs \cite{Cheng-2024-autodetect}.
Second, how to validate the evaluation results of superhuman models? Since human references may not work for judging the evaluation results of superhuman models, it is important to avoid the misleading evaluation results (like reward hacking \cite{rewardhacking} in RLHF) causing misaligned with human values \cite{mislead}.
Finally, today's evaluation is heavily reliable on static evaluation (i.e., results on benchmarks), but how can we design auto-evaluation methods for superhuman models and how can we conduct adversary tests automatically?






\section{A Framework to Realize Superalignment} 

\begin{figure}[htbp]
\centering
\includegraphics[scale=0.5]{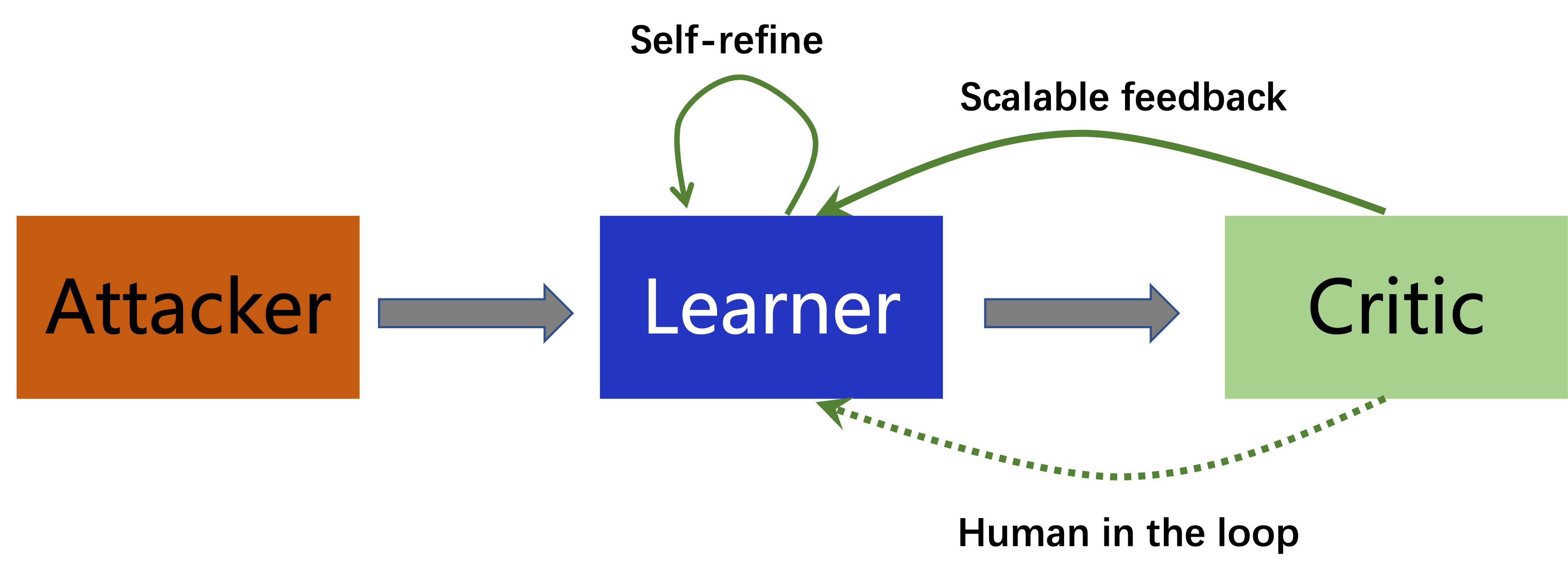}
\caption{A conceptual framework for superalignment. \textit{Attacker} generates adversary queries so that Learner may fail to produce high-quality responses; \textit{Learner}  will be continuously improved by learning from scalable feedbacks generated from Critic or from minimal human feedbacks whenever necessary; and \textit{Critic} generates explanations, feedbacks, or reasons given a query from the attacker and a response from the learner as input. Starting from some seed input, the pipeline can be automatically iterated.}
\label{fig:framework}
\end{figure}

In this section, we will present a feasible framework to realize superalignment, as presented in Figure \ref{fig:framework}. There
are three modules in this framework: \textit{an attacker model}, which simulates attacks and generates adversary queries such that a learner model may fail to produce high-quality responses; \textit{a learner model} which will be continuously improved by learning from scalable feedbacks generated from a critic model or human feedback whenever human intervention is required; and \textit{a critic model} which generates explanations, feedbacks, or reasons given a query from the attacker and a response from the learner as input. This pipeline can be automatically executed and iterated when it is started from some seed input. Noticeably, the attacker, learner, and critic can be the same foundation model but with different versions. 

This is a conceptual framework, which leaves many questions unsolved in implementation. In general purpose, it is very difficult to make the pipeline work smoothly, however, we believe in some specific cases, for instance, mathematic tasks and code generation, this framework is feasible and there are already some research attempts as shown in \cite{Cheng-2024-autodetect} and \cite{jiale-spar}.
In what follows, we will discuss the key challenges and research problems in this framework.

\subsection{Attacker: Discovering the weaknesses of LLMs automatically}
The attack model aims to generate adversary queries that the learner may fail to answer. In this manner, the weaknesses of the learner model can be automatically exposed, and then these weaknesses can be fixed accordingly. Such adversary attacks have been largely studied (known as red teaming methods) in safety issues of generative models \cite{Zou-2023-gcg}, image classification \cite{Liu-2024-image-rebustness}, or image generation in diffusion models \cite{Zhang-2024-attack-diffusion}.

However, building such an attack model has never been easy. One straightforward way is to use prompt engineering which designs some prompt templates to trigger a model to generate adversary attacks. Unfortunately, this method is sensitive to pre-specified prompts, largely depends on the base capability of the attack model, and may fail in some cases such as LLM's safety since many LLMs have been trained not to generate harmful queries. Another way is to train an attack model to simulate adversary attacks by constructing adversary training data. This can be enhanced by reinforcement learning. For instance, in the context of LLM's safety, Wen et al. \cite{Wen-2023-implicit} present a RL method for generating implicitly toxic contents with a reward function, which encourages the model to generate subtle, implicitly toxic contents. By this means, the generated contents have very high attack success rates to common toxicity classifiers. In general purpose, we can train the attack model with reinforcement learning, using the reward signal from a critic model, while the objective is to encourage the attack model to generate queries that lead to down-rated responses from a strong model.  Besides, the attacker could also red-team  LLM flaws
beyond safety issues. For example, Cheng et al.~\cite{Cheng-2024-autodetect} proposes a unified framework ``AutoDetect'', where three LLM-powered agents work collaboratively to automatically detect potential weaknesses in general-purpose tasks, such as mathematics and coding. 

\subsection{Learner: Learning from scalable feedbacks}

The learner model, which is the target model to be optimized in this framework, will make self-refinement by learning from scalable feedbacks from a critic model and also human feedbacks whenever necessary. The core of the learner model is alignment algorithms which enable the model to learn from scalable feedback. In past years, there have been some notable algorithms for this purpose: Proximal Policy Optimization (PPO) \cite{John-2017-ppo}, Direct Preference Optimization (DPO) \cite{Rafael-2023-dpo}, Efficient and Exact Policy Optimization (EXO)\cite{ji-2024-exo}, and many more. Since we have human experts in the loop, there raises a critical research problem in the learning process: how can the learner learn from mixed feedback signals, most times from model-generated feedback and rarely from human experts? In superalignment, designing more efficient and effective alignment algorithms is still the major challenge.

The form of feedback mainly falls into two types: a reward function which is trained on pair-wise preference data, or textual critics generated from a critic model. In our framework, we are more interested in textual critics as feedback since it does not require additional training to obtain a reward function. There are some critical research problems in this form of feedback: how can the learner learn from such textual critics? And which form of critics will be easier for the learner to learn from? Very recently, critic models such as criticGPT \cite{Nat-2024-criticGPT} and CritiqueLLM\cite{Ke-2024-Critique} have been proposed to generate scalable critics for diverse generation tasks from different dimensions, 
and with the assistance of such critic models, human experts are easier to provide reliable supervision,
however, how such critics can be used to improve the learner model is still an under-explored problem.

A portion of previous work \cite{RELC, beyond} focuses on employing critiques to facilitate more accurate and fine-grained reward estimation, thereby improving the performance of learners in an indirect way.
For instance, RELC\cite{RELC} utilizes critiques to decompose sequence-level rewards into segment-level ones, aiming to alleviate the issue of reward sparsity in PPO optimization. 
 Another line of research directly leverages critiques to refine the generated response through a refinement model. DRC\cite{drc} and FENCE\cite{fence} demonstrate that fine-grained critiques and refinements are more effective for enhancing the factuality of responses. 
In summary, due to its unique informational advantage over scalar rewards, natural language feedback holds significant potential for model optimization. However, what are the most efficient and learnable forms of critique is still largely under-explored.

\subsection{Critic: Generating scalable, faithful, and learnable critics} 

The critic model aims to generate scalable feedbacks for the learner model. The feedback, in the form of textual description in this paper, may be explanations or reasons why a response to an adversary query is good or bad. There are critical questions in building such critic model. 
First of all, the central role of the critic model lies in its criticizing ability: can the model generate relevant, informative, and discriminative explanations or reasons for a given query-response pair? 
Second, how can we ensure and evaluate the faithfulness of a generated critic? This problem is closely related to self-evaluation of a model-generated result \cite{llmjudge, panickssery2024llm}, probability calibration \cite{gao-etal-2024-bayesian}, and confidence estimation \cite{jung2024trust}. 
Third, how can the critic model generate critics that will be easily used to optimize the learner model? Such a critic will be called a learnable critic in this paper.

It is not trivial to train a general-purpose critic model that is generalizable across different generation tasks, topics, and evaluation dimensions. To evaluate various generation tasks, GPT-4 has been widely used to generate evaluative critics by prompt engineering. 
However, this method is faced with high cost, low stability, and low reproducibility. Moreover, the evaluation performance is largely determined by the ability of base models.
Therefore, training a specialized critic model has become a common choice \cite{autoj, judgelm} recently, with the aim of avoiding potential risks of commercial APIs, such as high cost, unstable usage, and data leakage.
However, it still faces challenges such as generalization capabilities and hallucinations, which hinder its further applicability.
Moreover, several works attempt to effectively utilize critiques through a new human-in-the-loop approach. 
OpenAI endeavors to train critic models to generate critiques in summarization \cite{self-critique} and code generation \cite{critiquegpt}, assisting human annotators in identifying mistakes in responses more easily.  
The results indicate that critiques not only enhance the coverage and accuracy of human annotators in detecting mistakes, but also help generative models refine their own answers to improve their quality further, 
demonstrating the significant potential of the critic model in research on scalable oversight.

\subsection{Realization of the superalignment framework}
We propose a conceptual framework for superalignment in previous sections and we highlight some key research problems in each module. We believe when these problems have been solved, it will be feasible to run the pipeline smoothly.  The essence of the superalignment framework lies in self-refinement or self-improvement: the learner can learn automatically and improve itself from scalable feedback.

Interestingly, there have been some notable research attempts similar to the idea of our proposed framework. These works are highly related to keywords such as bootstrapping, self-alignment, self-play, self-refine, etc. In 
the ``Self-Taught Reasoner" (STaR) \cite{zelikman2022star}, a \textit{bootstrapping} reasoning technique was proposed. This method
relies on a simple loop: generate rationales to answer questions by prompting the model with a few rationale examples; if the generated answers are wrong, try again to generate a rationale given the correct answer; ﬁne-tune the model on all the rationales that ultimately yielded correct answers; then iterate the process.
 A \textit{self-alignment} framework was proposed by Yuan et al. \cite{yuanself}, which consists of two steps: at the self-instruction creation step, some newly created prompts are used to generate candidate responses from an earlier version model $M_t$, which also predicts its own rewards via LLM-as-a-Judge prompting;  at the instruction following training step, preference pairs are selected from the generated data using the reward signals, on which a new model $M_{t+1}$ was trained using DPO algorithm\cite{Rafael-2023-dpo}. The procedure can be iterated, leading to not only improved instruction following but also stronger reward modeling ability. 
Similarly, this idea was explored in SPIN (Self-Play fIne-tuNing)\cite{chenself}: starting from an SFT dataset and an initial model $M_0$, the method generates synthetic data from an earlier model $M_t$, and then train a new version $M_{t+1}$ using DPO algorithm; in the next iteration, the new version $M_{t+1}$ is treated as a supervised fine-tuning model to obtain a newer one $M_{t+2}$. Unlike self-alignment which selects preference data using self-rewarding signals, SPIN assumes model-generated data are always worse than the human data in the SFT dataset when constructing preference pairs. 
Their results show that SPIN can convert a weaker LLM to a stronger LLM 
and thereby demonstrate the promise of self-play.
Another interesting idea, which is largely explored in the community, is \textit{self-refine}~\cite{madaan2024self}: 
an LLM first generates an initial output and then provides feedback for its output, and then the model uses the feedback to refine its output; the process can be repeated iteratively. Self-Refine uses a single LLM as the generator, refiner, and feedback provider, and requires no additional training.
Cheng et al. \cite{jiale-spar} propose a self-refinement framework, SPAR, which involves an actor model to be optimized and a refiner model that critiques and generates improved responses through tree-search sampling. This framework effectively scales inference-time computation to construct high-quality training data, enabling continuous self-improvement for both the actor and the refiner through iterative training.

Some other works attempt to identify the weaknesses in the system automatically and fix them accordingly.
Cheng et al. \cite{Cheng-2024-autodetect} introduce AutoDetect, a framework designed to automatically identify weaknesses in LLMs across various tasks. AutoDetect features three agents: Examiner, which creates a detailed task taxonomy; Questioner, which generates queries; and Assessor, which analyzes low-scored cases to propose potential weaknesses. The Questioner’s queries are input to a target model, and responses are scored to identify weak points. This framework has achieved a success rate of over 30\% in top models like ChatGPT and Claude. Additionally, the identified weaknesses can help enhance models, such as the LLaMA series, through supervised fine-tuning.
Bai et al. \cite{bai2024benchmarking} propose the Language-Model-as-an-Examiner framework, designed to automatically benchmark the knowledge of foundation models. This framework employs an LLM as an examiner to generate diverse questions across domains, probe deeper knowledge through follow-up queries, and evaluate the model's responses. Beyond assessing performance, this approach can also serve as a tool for identifying knowledge-related weaknesses in the tested models.
Cohen et al. \cite{cohen2023lm} introduce a cross-examination-based framework for evaluating the factuality of language models. This approach involves two interacting LMs: the Examinee, which generates claims, and the Examiner, which conducts a multi-turn interaction to identify inconsistencies in the Examinee's responses. Inspired by legal truth-seeking mechanisms, the Examiner crafts targeted questions to uncover contradictions and expose factual inaccuracies in the Examinee's claims.


Despite impressive empirical progresses, a fundamental understanding of LLM self-improvement remains very limited, thereby requiring much more in-depth theoretical modeling and empirical analysis. Some works reported that use of model-generated data to next-generation model recursively can lead to \textit{model collapse}\cite{shumailov2024curserecursiontraininggenerated}: a degenerative learning process where models start forgetting improbable events over time, as the models become poisoned with its own generated, biased data. In image generation, \cite{alemohammad2023selfconsuminggenerativemodelsmad} showed without enough fresh real data in each generation of an autophagous loop, future generative models can have progressive decrease in output quality or diversity. In \cite{guo2024curiousdeclinelinguisticdiversity}, the authors discovered a consistent decrease in the diversity of model outputs through iterative training, particularly for those highly creative tasks, thereby underscoring the potential risks of training language models on synthetic text, particularly regarding the preservation of linguistic richness. Similarly, \cite{briesch2024largelanguagemodelssuffer} showed the
self-refinement training loop can lead to declines in output diversity depending on the proportion of the used generated data and fresh data can slow down this decline, but not stop it. In \cite{wu2024progressregressselfimprovementreversal}, the authors also observed declines in output diversity and out-of-distribution (OOD) generalization during LLM self-refinement training. 
Most interestingly, \cite{song2024mindgapexaminingselfimprovement} presents a mathematical formulation for self-improvement and formalizes a concept of \textit{generation-verification gap}, and the authors reveal that the gap between the verification capability (judging the quality of generations) and the generation capability is the force to drive self-refinement. They studied verification mechanisms to improve self-refinement, for instance, an ensemble of different verification methods can enhance self-improvement. We believe this is the most in-depth theoretical analysis on self-refinement up to now.

\section{Conclusion and Future Directions}
In this paper we discuss the superalignment of superhuman AI systems with large language models. We give an informal definition of superalignment by outlining the shift of learning paradigms from pretraining, supervised finetuning, alignment, and superalignment. Afterwards, we highlight some key research problems in superalignment including weak-to-strong generalization, scalable oversight, and evaluation of the alignment. Then, we present a conceptual framework to realized superalignment, which consists of three components: attacker which aims to discover the weaknesses of LLMs automatically, learner which learns from scalable feedbacks (mixture of AI and human feedbacks), and critic which produces scalable, faithful and learnable critics. We highlight some critical research problems in each component, and also summarize some major research advancements in these sub-directions. Finally, we also summarize some interesting research attempts that are highly related or may lay a foundation to superalignment: self-alignment, self-play, self-refinement, and others. These works can be viewed as early attempts towards superalignment, and show promising results, thereby partially verifying the feasibility of the framework proposed in this paper. 

Though still in its infancy, superalignment poses new research problems that is worthy to study in near future: 

\textbf{Identifying new emergent risks of superhuman intelligence}: The safety of superhuman AI systems has gained much attention recent years, and many safety issues have been revealed and studied, including discrimination, bias, property and privacy, misinformation and disinformation, ethics, social norms, and many more. We call such safety issues \textit{low-order} safety problems as the harms can be directly recognized by superficial clues shown in the generated content. However, \textit{high-order} safety problems such purposely deception to mislead human and manipulation of human beliefs may be more subtle, indirect, and complicated to identify, and require long-term evaluation. 
Moreover, unknown risks in specialized domains (eg. biological threats) are also very dangerous threats to our society. How to recognize, identify and evaluate such unknown risks in high-stake fields is very critical to AI safety.

\textbf{Providing reliable and scalable oversight to superhuman models}: We have discussed some works on self-alignment, self-play and self-refinement, where these works share in common that a model is iteratively refined with synthetic data. However, in superalignment, how to synthesize high-quality data that the current model is not good at modeling is challenging, and how can we provide reliable oversight on such synthetic data is largely requiring human-AI collaboration. Due to the scalability issue, at most time we have to rely on AI feedbacks, but when human experts will intervene and how they will be evolved in the pipeline is a complex problem. There are still many research problems worth to do in the future.


\textbf{Aligning large language models from multiple dimensions:} Aligning large language models to human values is an extremely complex problem and requires to consider very diverse aspects, cultures, regions and countries. Existing works mainly focus on single perspective, however, modeling very different perspectives such as values, safety, social norms, and ethics in one paradigm is yet to be consider. Therefore, it is indispensable to design multi-objective optimization alignment algorithms to model these factors simultaneously \cite{zhong2024panacea}. 
This is quite challenging since these social perspectives are interleaved together and have different meanings in different contexts (such as cultures, countries or regions).


\newpage

\Acknowledgements{This work was supported by the National Natural Science Foundation of China (Distinguished Young Scholar Project with Grant No. 62125604) and key project (Grant No. 61936010). We would like to thank student Bosi Wen and Jiale Cheng for minor edits of the manuscript, and thank the students for earlier discussions of this superalignment framework, including Chujie Zheng, Haozhe Ji, Yuxian Gu, Zhexin Zhang, and many others.}



\bibliographystyle{unsrt} 
\bibliography{reference}

\end{document}


\ArticleType{Supplementary File}

\title{Title}{Title for citation}

\author[1]{Aaa AUTHOR}{}
\author[1,2]{Bbb AUTHOR}{{bauthor@xxx.com}}
\author[2]{Ccc AUTHOR}{}
\author[3]{Ddd AUTHOR}{}


\address[1]{Affiliation, City 000000, Country}
\address[2]{Affiliation, City 000000, Country}
\address[3]{Affiliation, City 000000, Country}

\maketitle


\begin{appendix}

\section{Importance}
Please use this sample as a guide for preparing your letter. Please read all of the following manuscript preparation instructions carefully and in their entirety. The manuscript must be in good scientific American English; this is the author's responsibility. All files will be submitted through our online electronic submission system at \href{https://mc03.manuscriptcentral.com/scis}{HERE}.

\section{More information}
The examples at the bottom of the .tex file can help you when preparing your manuscript. We are appreciate your effort to follow our style~\cite{1,2}.

\end{appendix}
